\documentclass[letterpaper, 10 pt, conference]{ieeeconf}  

\IEEEoverridecommandlockouts                              

\usepackage{amssymb}  

\title{\LARGE \bf
Design and Visual Servoing Control of a Hybrid Dual-Segment Flexible Neurosurgical Robot for Intraventricular Biopsy
}

\author{Jian Chen, Mingcong Chen, Qingxiang Zhao, Shuai Wang, Yihe Wang, Ying Xiao, Jian Hu, \\ \enskip \enskip Danny Tat Ming Chan, Kam Tong Leo Yeung, David Yuen Chung Chan and Hongbin Liu 
\thanks{Jian Chen is with the School of Artificial Intelligence, University of Chinese Academy of Sciences, Beijing 100049, China, also with the State Key Laboratory of Multimodal Artificial Intelligence Systems, Institute of Automation, Chinese Academy of Sciences (CASIA), Beijing 100190, China, and also with the Centre of AI and Robotics, Hong Kong Institute of Science and Innovation, Chinese Academy of Sciences (CAIR-HKISI-CAS), HKSAR {\tt\small chenjian2020@ia.ac.cn}}
\thanks{Mingcong Chen is with Department of Biomedical Engineering, City University of Hong Kong, HKSAR {\tt\small mingcong.chen@my.cityu.edu.hk}}
\thanks{Qingxiang Zhao, Shuai Wang, Jian Hu and Yihe Wang are with CAIR-HKISI-CAS, HKSAR {\tt\small qingxiang.zhao, shuai.wang, hujian, Yihe.Wang@cair-cas.org.hk}}%
\thanks{Ying Xiao is with the State Key Laboratory of Multimodal Artificial Intelligence Systems, CASIA, Beijing 100190, China {\tt\small ying.xiao@ia.ac.cn}}%
\thanks{Danny Tat Ming Chan, Kam Tong Leo Yeung, David Yuen Chung Chan are with Department of Surgery, The Chinese University of Hong Kong, HKSAR {\tt\small tmdanny, leoyeung, davidchan@surgery.cuhk.edu.hk}}%
\thanks{Hongbin Liu is with the State Key Laboratory of Management and Control for Complex Systems, CASIA, Beijing 100190, China, also with CAIR-HKISI-CAS, HKSAR, and also with the School of Biomedical Engineering and Imaging Sciences, King's College London, London SE1 7EU,UK {\tt\small liuhongbin@ia.ac.cn}}
\thanks{Corresponding author: Hongbin Liu.}
}

\usepackage{amsmath}
\usepackage{float}
\usepackage{graphicx}
\usepackage{subfigure} 
\usepackage{svg}
\usepackage{bm}
\usepackage[T1]{fontenc}
\usepackage{geometry}
\usepackage[colorlinks, linkcolor=blue, citecolor=red]{hyperref}
\begin{document}
\maketitle
\thispagestyle{empty}
\pagestyle{empty}

\begin{abstract}
Traditional rigid endoscopes have challenges in flexibly treating tumors located deep in the brain, and low operability and fixed viewing angles limit its development. This study introduces a novel dual-segment flexible robotic endoscope MicroNeuro, designed to perform biopsies with dexterous surgical manipulation deep in the brain. Taking into account the uncertainty of the control model, an image-based visual servoing with online robot Jacobian estimation has been implemented to enhance motion accuracy. Furthermore, the application of model predictive control with constraints significantly bolsters the flexible robot's ability to adaptively track mobile objects and resist external interference. Experimental results underscore that the proposed control system enhances motion stability and precision. Phantom testing substantiates its considerable potential for deployment in neurosurgery.
\end{abstract}

   
\section{Introduction}
\begin{figure}[t]
      \centering
        \includegraphics[width=0.48\textwidth]{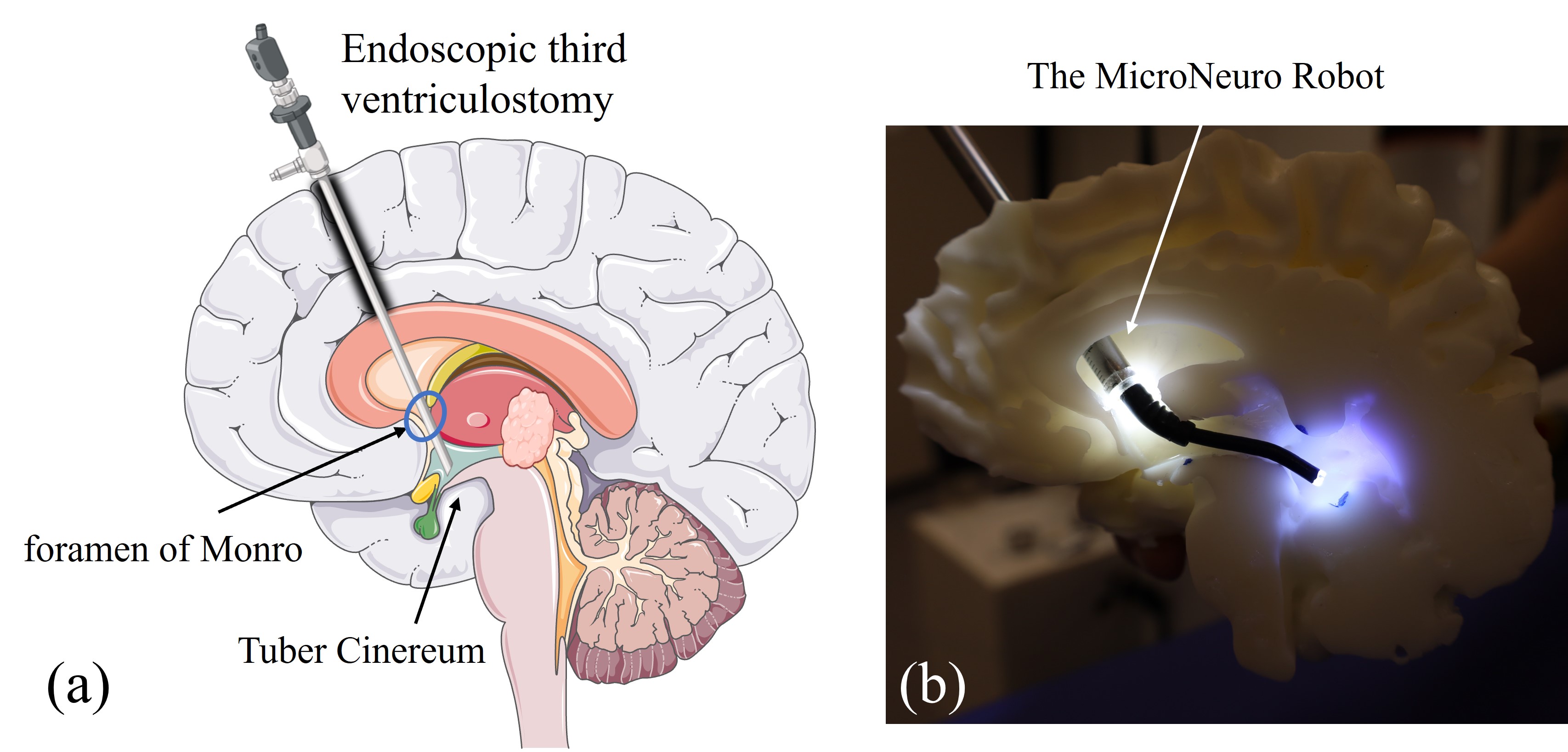}
      \caption{(a) Traditional rigid intraventricular endoscopes can only move forward and backward along the axis. (b) The MicroNeuro flexible robot system reach with one burr hole.}
        \label{fig1}
         \vspace{-2mm}
   \end{figure}

Tumors located within the brain's ventricular system pose significant health risks and present considerable treatment challenges due to their difficult-to-reach locations and proximity to critical neurological structures \cite{yacsargil2008surgery}. Over the past three decades, rigid endoscopes have emerged as the primary tool for visualization in diverse intraventricular neurosurgical procedures \cite{rigante2019overview}. For instance, the MINOP endoscope (Aesculap Inc., PA, USA) is employed for intraventricular indications, while the LOTTA endoscope (Karl Storz SE \& Co.KG,Tuttlingen, Germany) is preferred for patients with small ventricles. Unfortunately, conventional neurosurgery with rigid endoscopes still has two primary limitations: (i) The rigid structure limited maneuverability \cite{rigante2019overview} within the complex anatomy of the brain, slight movement abruptly or incorrectly may lead to potential brain trauma and complications; and (ii) the limitation of fixed viewing angles of rigid instruments, complicating the biopsy of tumors in difficult locations, as shown in Fig. \ref{fig1}(a). While flexible robots can enhance endoscope dexterity, their use has been limited by the lower-resolution visualization \cite{chowdhry2013intraventricular}, poor accessibility of single flexible 
segment on traditional endoscopes and the procedural complexities of combined rigid and flexible endoscopy \cite{amer2018combined}. The confined intracranial space also demands high dexterity and compliance from flexible surgical tools \cite{zeng2021modeling, qi2021toward}, presenting additional control challenges \cite{yoon2013compact}.

With the real-time visual feedback from the robot tip, image-based visual servoing (IBVS) is particularly compatible with this eye-in-hand configuration \cite{chen2023fully}. The classical IBVS has been widely used to solve the tracking \cite{li2023towards}, shape control \cite{xu2022adaptive}, depth estimation \cite{fallah2020depth} problems of flexible endoscopes. During neurosurgical endoscopic operations, external interference, such as inserting internal instruments, may lead to potential issues with the proportional controller. These issues could manifest as slow convergence \cite{nazari2022visual} and decreased tracking performance \cite{li2023towards}. To enhance the robustness, Jiang et al. \cite{jiang2022robust} combined a sliding mode control (SMC) with IBVS to overcome the system uncertainties. For environmental interaction, Oliva et al. \cite{oliva2022towards} presented a dynamic IBVS controller with an Extended Kalman Filter (EKF) to improve the tracking speed and stability.

However, most of the above mentioned methods did not take surrounding constraints into account, which is indispensable in neurosurgery. During intraventricular biopsies, unconstrained movement may damage significant nerves or blood vessels \cite{azab2014overview}. Model predictive control (MPC) \cite{book,lin2019multi} utilizes constraints to ensure control actions and system states remain within desired bounds throughout the control horizon. A MPC framework within a visual servoing scheme was proposed to achieve precision manipulation in \cite{calli2017vision} to deal with the model inaccuracies. Notably, the inherent robustness characteristics of IBVS and MPC significantly improve controller performance  \cite{bechlioulis2019robust}. Chen et al. \cite{chen2022qpso} utilized a QPSO-MPC based tracking method for a continuum robot arm. Chien et al. \cite{chien2021kinematic} also used MPC method to control the position of a continuum robot, where the inverse kinematics was estimated as the basis. Therefore, the complex model transfer chain could be represented by Jacobian and the surroundings obtained by endoscopic camera passes constraints into MPC control scheme, which are applicable for MIS-oriented scenarios for continuum robots.

To address the design and control issues mentioned above, this work makes two main contributions: (i) a cable-driven hybrid dual-segment flexible endoscope for the intraventricular neurosurgery is proposed, which could pass through one single burr hole and provides sufficient dexterity to biopsy in the narrow ventricle, as shown in Fig. \ref{fig1}(b); (ii) a visual model predictive control framework with the online Jacobian estimation is proposed to enhance the robustness of visual servoing control. The rest of this work is organized as follows. Section II details design rules and prototype. In Section III, the kinematics model of the robot and camera is established with an online Jacobian estimation. Besides, Section IV introduces the visual MPC algorithm. Section V illustrates the effectiveness of the robot and the proposed methods. Finally, Section VI concludes this work.

\section{Mechanical Design}
\subsection{Design Goals}
The MicroNeuro is designed for intraventricular neurosurgery. Based on knowledge of brain anatomy and clinical demand from surgeons, the main design goals are first summarized as follows:

\begin{enumerate}
    \item Dimension: The mean diameters of the foramen of Monro (FM) were 5.7 mm on the axial image, 7.8 mm on the coronal image, and 5.6 mm on the sagittal image \cite{zhu2013single}.  Thus, the outer diameter of the flexible endoscope should be less than 5.4 mm to avoid collision with the FM.
    \item Endoscope features: The MicroNeuro should provide high quality images and a working channel for biopsy instruments. Since clinical surgery needs to be performed underwater, the MicroNeuro also needs to provide irrigation and suction functions.
    \item Dexterity: Deflective length of the MicroNeuro should be short and able to bend with a large curvature.
\end{enumerate}

  \begin{figure}[!t]
     \vspace{3mm}
      \centering
        \includegraphics[width=0.48\textwidth]{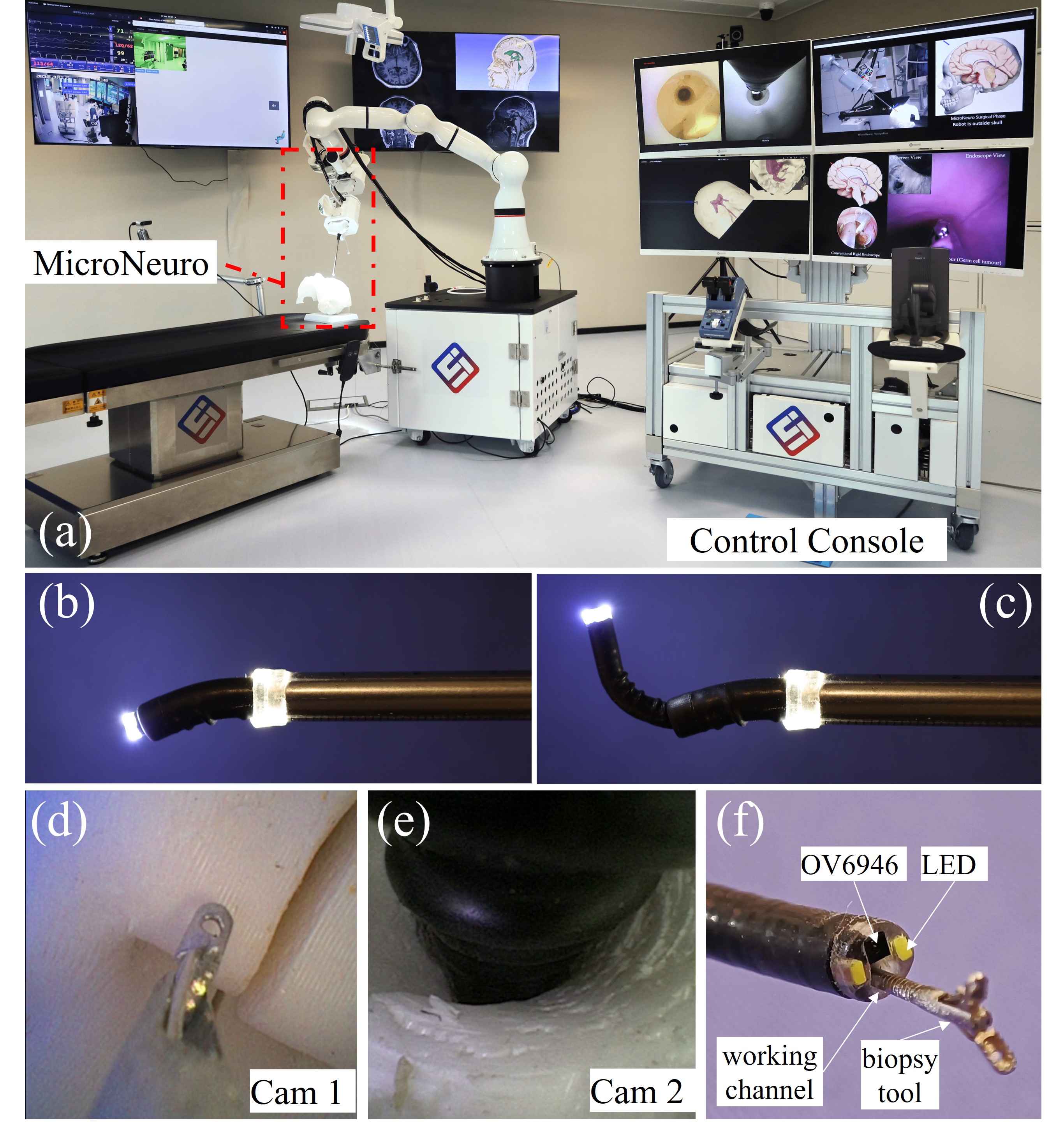}
      \caption{Overview of The robot system. (a) The MicroNeuro system and MicroNeuro surgical robot. (b) Steering mode 1 without insertion of the distal. (c) Steering mode 2 with $S$ shape. (d) Endoscopic view of inner endscope. (e) Endoscopic view of outer sheath. (f) Endoscope features.}
        \label{fig:wholeview}
         \vspace{-2mm}
   \end{figure}

\subsection{System Overview}
This work was developed based on the surgical robot system for neurosurgery, designated  MicroNeuro \cite{microneuro}. As shown in Fig. \ref{fig:wholeview}(a), this system mainly consists of the MicroNeuro and its actuation units, which are mounted on the end of a 7 DoFs robot arm (ER7 Pro, ROKAE). The quick release mechanism of the MicroNeuro facilitates the individual disinfection of endoscopes. Besides, a control console is also built for master-slave teleoperation with four monitors, a foot pedal, a joystick (TCA, THRUSTMASTER) and a master device (TouchX, 3D SYSTEM). 

The MicroNeuro consists of two bendable flexible robots which are connected to a rigid tube. As shown in Fig. \ref{fig:wholeview}(d), (e) and (f), it provides several functions, such as multi-view images,  water irrigation and suction, working channel (diameter 1.2mm) and illumination. The distal end of the inner endoscope and the rigid catheter are each equipped with a camera (OV6946). Unlike conventional dual-segment flexible robots with fixed length, each robot of the MicroNeuro can be axially translated relative to each other, so two combined bending modes can be realized: (i) mode 1 [see Fig. \ref{fig:wholeview}(b)], the inner endoscope has no axial movement, and only the outer flexible sheath bends; (ii) mode 2 [see Fig. \ref{fig:wholeview}(c)], the inner endoscope could be inserted independently (maximum distance is 40mm).

\subsection{Hybrid Dual-Segment Flexible Endoscope Design}
\begin{figure}[t]
     \vspace{3mm}
      \centering
        \includegraphics[width=0.48\textwidth]{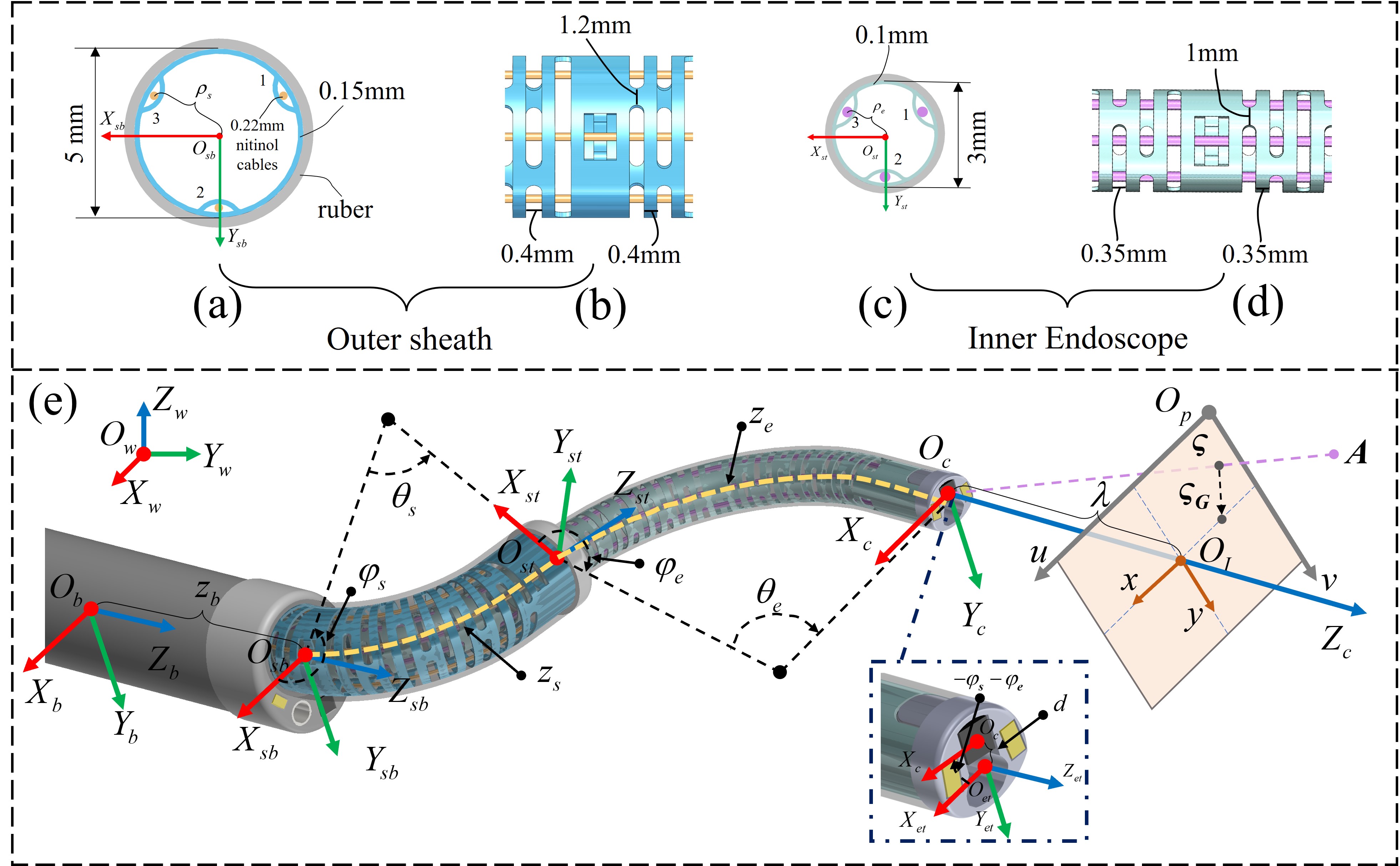}
      \caption{Mechanical design of the MicroNeuro robot. (a) Axial section view of outer sheath with cables distribution diagram. (b) Nitinol backbone of the outer sheath. (c) Axial section view of inner endoscope. (d) Nitinol backbone of the inner endoscope. (e) Illustration of coordinate frames.}
        \label{fig:flexible}
         \vspace{-2mm}
   \end{figure}

The backbones of each flexible robot are manufactured by femtosecond laser cutting of superelastic nitinol tubes. Fig. \ref{fig:flexible} shows the parameter definitions and values. The two robots have multiple pairs of notched joints distributed along the axial direction, and each joint has a bidirectional symmetrical rectangular notch. Three nitinol cables, driven by brushless coreless motors (ASSUN), are welded to the distal end of each flexible robot and routed along a crimped grooves.

\section{Modelling}
\subsection{Kinematics of MicroNeuro}
The distribution of notches in the backbone makes it axial stiffness larger than that in lateral direction, so the backbone would bend when the eccentrically fixed cables are stretched. Referring to the piecewise constant curvature (PCC) model \cite{webster2010design}, each segment of MicroNeuro bends with a constant curvature along its length, similar to a circular arc, when actuated. As shown in Fig. \ref{fig:flexible}, MicroNeuro can be geometrically parameterized by $\bm{{\Phi}}=\left(\begin{matrix}z_b&\theta_s&\varphi_s&z_e&\theta_e&\varphi_e\end{matrix}\right)^{\mathrm{T}}$ in the configuration space, where ${z_b}$ is the overall insertion distance provided by the robot arm, $\theta_s$ and $\theta_e$ are the bending angles, $\varphi_s$  and $\varphi_e$ are the rotation angles between the bending plane and the $oxz$ plane, and $z_e$ is the variable length of the inner endoscope, provided by the servo motors. $\theta_s$, $\varphi_s$, $\theta_e$, $\varphi_e$ can be calculated from the actuator space variables $\bf q=\left(\begin{matrix}z_b&l_{s,1}&l_{s,2}&l_{s,3}&z_e&l_{e,1}&l_{e,2}&l_{e,3}\end{matrix}\right)^{\mathrm{T}}$: 
\begin{equation}
\label{eq:angle_param}
\begin{split}
{\theta_i}&={\frac{2\sqrt{l^2_{i,1}+l^2_{i,2}+l^2_{i,3}-l_{i1}l_{i2}-l_{i2}l_{i3}-l_{i1}l_{i3}}}{3\rho_i}}\\
{\varphi_i}&=\mathrm{tan}2({{l_{i1}+l_{i3}-2l_{i2}},{\sqrt{3}(l_{i3}-l_{i1})}})
\end{split}
\end{equation}
where $i\in\left\{e,s\right\}$, and the subscripts $e$ and $s$ used to represent the outer sheath and inner endoscope, respectively, $\rho_i$ is the distance between the center of the cable and the center of the robot, $l_{i,m}$, $m\in\left\{1,2,3\right\}$ are the length of the driving guide wires in each flexible robot. The transformation matrix  ${^{b}_{{et}}}{{\mathrm {\bf T}}}\in{\mathbb{R}}^{4\times4}$ from the base frame $O_bX_bY_bZ_b$ to the robot tip frame $O_{et}X_{et}Y_{et}Z_{et}$ is:
\begin{equation}
\label{eq:base2tip}
\begin{split}
{}_{{\bf{et}}}^{\bf{b}}{\bf{T}} = &{{\bf{\tau }}_{\bf{z}}}({z_b}){{\bf{R}}_{\bf{z}}}({\varphi _s}){{\bf{\tau }}_{\bf{x}}}(\frac{{{z_s}}}{{{\theta _s}}}){{\bf{R}}_{\bf{y}}}({\theta _s}){{\bf{\tau }}_{\bf{x}}}( - \frac{{{z_s}}}{{{\theta _s}}})\\
&{{\bf{R}}_{\bf{z}}}({\varphi _e}){{\bf{\tau }}_{\bf{x}}}(\frac{{{z_e}}}{{{\theta _e}}}){{\bf{R}}_{\bf{y}}}({\theta _e}){{\bf{\tau }}_{\bf{x}}}( - \frac{{{z_e}}}{{{\theta _e}}})
\end{split}
\end{equation}
where ${{\bf{\tau }}_{\bf{j}}}$, ${\bf R_j}\in{\mathbb{R}}^{4\times4}$  respectively denote translation and rotation about axis j, $z_s$ is the length of the outer robot. Considering the offset $d$ of the camera frame $O_cX_cY_cZ_c$ from the robot tip, the camera w.r.t. the base is
\begin{equation}
\label{eq:cam2tip}
\begin{split}
{}_{{\bf{c}}}^{\bf{b}}{\bf{T}} = {}_{{\bf{et}}}^{\bf{b}}{\bf{T}}{{\bf{R}}_{\bf{z}}}(-{\varphi _s}-{\varphi _e}){{\bf{\tau }}_{\bf{y}}}(-d)
\end{split}
\end{equation} 


The Jacobian matrix ${{\bf{J}}_{\bf{r}}}\in{\mathbb{R}}^{3\times6}$ is used to analytically establish the approximate relationship between camera velocity and joint velocity. Considering the translational motion, at discrete instance $k$, the iterative form is $\Delta{\bf{P_r}}(k)={\bf{J_r}}(k)\Delta{\bf{\Phi}}(k)$, where $\Delta{\bf{P}}(k)={\bf{P}}(k+1)-{\bf{P}}(k)$ is the small displacement of the camera, ${\bf{\Phi}}(k)={\bf{\Phi}}(k+1)-{\bf{\Phi}}(k)$ and ${\bf{J_r}}(k)$ can be derived through forward kinematics ${}_{{\bf{et}}}^{\bf{b}}{\bf{T}}$. To reach a given target position ${{\bf{P}}_{\bf{G}}}\in{\mathbb{R}}^{3}$ of the end of the robot in $O_{et}X_{et}Y_{et}Z_{et}$, we need to inversely solve the appropriate joint configuration. The damped least squares method \cite{buss2004introduction} provides an alternative Jacobian matrix to avoid joint velocity near singularities, i.e.
\begin{equation}
\label{eq:dls}
\begin{split}
\Delta{\bf{\Phi}}(k)={\bf{J^T_r}}(k)({\bf{J_r}}(k){\bf{J^T_r}}(k)+\sigma{I})^{-1}({\bf{P_G}}(k)-{\bf{P}}(k))
\end{split}
\end{equation}


\subsection{Visual Servoing Modeling}
However, material nonlinearity, segment interaction, external loads, etc. may have a significant negative impact on the accuracy of the PCC model. In this work, we consider a moving camera while the targets are fixed at any instance $k$. As shown in Fig. \ref{fig:flexible}(e), for a given point ${\bf{A}}\in\mathbb{R}^3$ in $O_cX_cY_cZ_c$, its coordinates in the image frame $O_Ixy$ and pixel frame $O_{p}uv$ are ${\bf{A}}(k)=(x,y)^{\mathrm{T}}$ and ${\bf{\varsigma}} (k)=(u,v)^{\mathrm{T}}$, respectively. According to the pinhole camera model, the perspective equation can be obtained from the relationship on similar triangles, i.e.
\begin{equation}
\label{eq:pinhole}
\begin{split}
u=\frac{\lambda_xx}{\lambda}+c_c,\enskip v=\frac{\lambda_yy}{\lambda}+c_y
\end{split}
\end{equation}
The motion of the features $\Delta\bm{\varsigma}(k)$ on the pixel plane can be predicted using the interaction matrix:
\begin{equation}
\label{eq:motionfeature}
\begin{split}
\Delta\bm{\varsigma}(k)={\bf{L_m}}(k)\Delta{\bf{P}}(k)
\end{split}
\end{equation}
where ${\bf{L_m}}\in\mathbb{R}^{2\times3}$ is a block matrix of ${\bf{L_o}}=[{\bf{L_m}}^{2\times3}|{\bf{L_\omega}}^{2\times3}]$ related to linear velocity, and
\begin{equation}
\label{eq:lo}
{\bf{L_o}}=\begin{bmatrix}-\frac{\lambda}{z_c} & 0 & \frac{x}{z_c}  & \frac{xy}{\lambda}  & -\frac{\lambda^2+x^2}{\lambda} & y\\  0 & -\frac{\lambda}{z_c} & \frac{y}{z_c}  & -\frac{\lambda^2+y^2}{\lambda} & -\frac{xy}{\lambda} & -x\end{bmatrix}
\end{equation}
where $\lambda_x$, $\lambda_y$ are the focal length in pixels, $c_x$, $c_y$ are optical center in pixels and $\lambda$ is focal length in millimeter. Define ${\bf{J_a}}(k) \in \mathbb{R}^{6\times8}$ as the Jacobian matrix between the actuator space and the configuration space from Eq. (\ref{eq:angle_param}), that is, $\Delta{\bf{\Phi}}(k)={\bf{J_a}}(k)\Delta{\bf{q}}(k)$. Combined Eq. (\ref{eq:dls}) and (\ref{eq:motionfeature}), the overall Jacobian matrix between pixel velocity and actuator velocity can be derived as follow:
\begin{equation}
\label{eq:jacobian}
\Delta\bm{\varsigma}(k)={\bf{L_m}}(k){\bf{J_r}}(k){\bf{J_a}}(k)\Delta{\bf{q}}(k)
\end{equation}

\subsection{Jacobian Matrix Estimation}
In the classic IBVS \cite{chaumette2006visual}, there are several choices for the depth $z_c$ in the matrix ${\bf{L_m}}(k)$. In this study, the depth $z_c^*$ at the desired position was used, and ${\bf{\hat{L_m}}}(k)$ denotes the estimation matrix. 

As a continuum robot, MicroNeuro has infinite DoFs. When subject to model mismatch problems caused by disturbance or manufacturing error, the model-dependent robot Jacobian matrix ${\bf{J}}(k)={\bf{J_r}}(k){\bf{J_a}}(k)$ may cause control deviations and need to be estimated online. First, the Jacobian estimate at $k=0$ needs to be obtained offline, then the Jacobian can be updated iteratively online during the robot movement.
\begin{enumerate}
    \item Initialization: A small actuator movement $\Delta{\bf{q_+}}(0)$ is imposed on the MicroNeuro while it is located outside the brain, and an external electromagnetic sensor (NDI Aurora) is mounted on the tip of MicroNeuro to measure the displacement. The $i$-th independent actuator variables $\Delta{{ q_{+,i}}}(0)$ causes a position deviation of the camera $\Delta{\bf{P_{c,i}}}(0)$. Hence, ${\bf{\hat{J_+}}}(0)$ is constructed as:
\begin{equation}
\label{eq:j+}
{\bf{\hat{J_+}}}(0)=\left[\begin{matrix}
\frac{\Delta{\bf{P_c,0}}(0)}{\Delta{{ q_{+,0}}}(0)}&\cdots&\frac{\Delta{\bf{P_c,8}}(0)}{\Delta {{q_{+,8}}}(0)}
\end{matrix}\right]
\end{equation}
To reduce manufacturing error, ${\bf{\hat{J_-}}}(0)$ is similarly constructed while a opposite displacement $\Delta {\bf{q_{-}}}(0)=-\Delta{\bf{q_+}}(0)$ is imposed. ${\bf{\hat{J}}}(0)$ is set as:
\begin{equation}
\label{eq:jhat0}
{\bf{\hat{J}}}(0)=0.5({\bf{\hat{J_+}}}(0)+{\bf{\hat{J_-}}}(0))
\end{equation}
    \item Online Estimation: The alterations in the MicroNeuro position and Jacobian matrix between adjacent instance are small, thus, the current analytical Jacobian matrix ${\bf{{J}}}(k)$ could be appropriately adjusted using ${\bf{\hat{J}}}(k-1)$:
    \begin{equation}
    \label{eq:analyticalJ}
    {\bf{\hat{J}}}(k)=(1-\omega(k)){\bf{J}}(k)+\omega(k){\bf{\hat{J}}}(k-1)
    \end{equation}
    where $\omega(k)=\frac{1}{1+\epsilon(k)}$ is the weighting factor, and $\epsilon(k)=||\bm{\varsigma}(k)-{{\bm{\varsigma}}_{\bf{G}}}(k)||_2$ denotes as the distance between measured feature $\bm{\varsigma}(k)$ and the target feature ${{\bm{\varsigma}}_{\bf{G}}}(k)$. The normalized $\bm{\varsigma}(k)$ and ${{\bm{\varsigma}}_{\bf{G}}}(k)$ could be applied in $\omega(k)$.
\end{enumerate}

  \begin{figure}[b]
      \centering
        \includegraphics[width=0.48\textwidth]{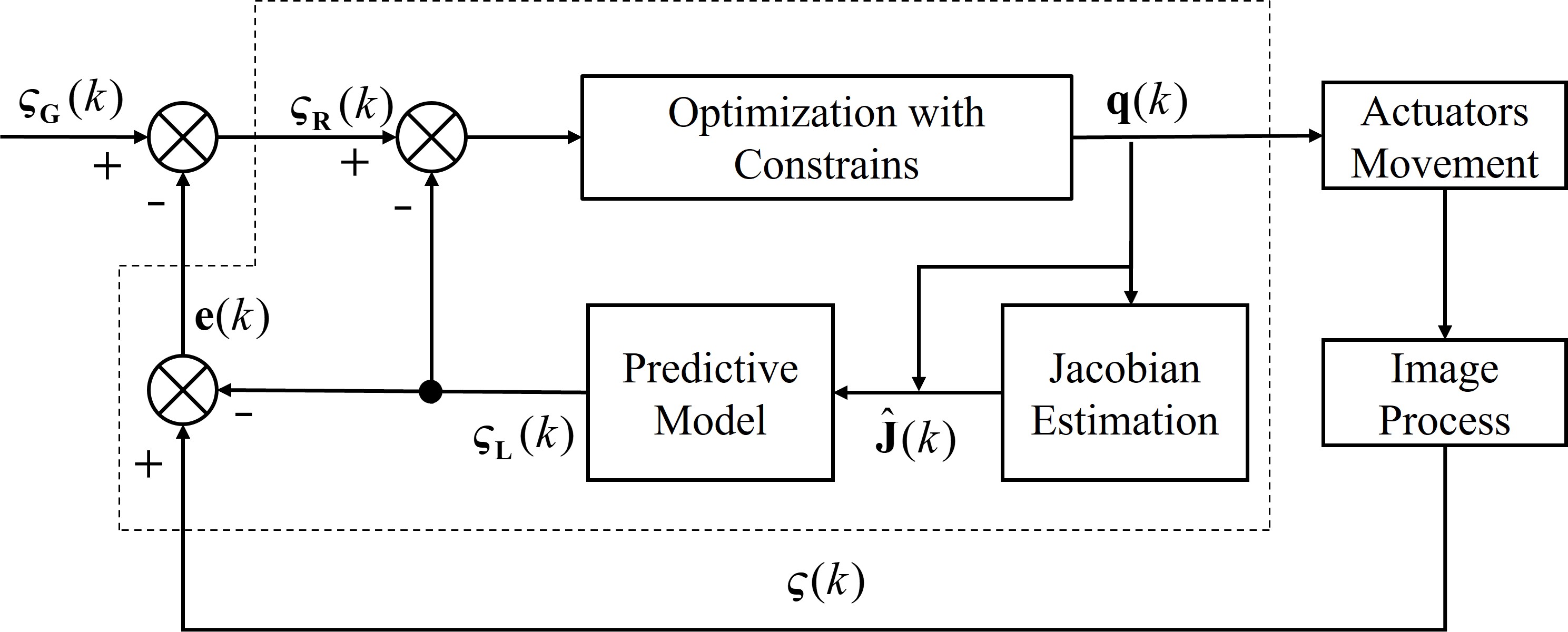}
      \caption{The visual MPC controller using an IMC scheme.}
      \label{fig:IMC}
         \vspace{-2mm}
   \end{figure}
   
\section{Visual Model Predictive Controller}
\subsection{Predictive Model}
The goal of the IBVS task is to minimize the error $\omega(k)$. Inspired from \cite{allibert2010predictive, norouzi2021constrained}, to reduce the negative impact of model inaccuracy and external disturbance, an internal model control (IMC) scheme \cite{saxena2012advances} is applied in the visual MPC controller, as shown in Fig. \ref{fig:IMC}. ${\bf{e}}(k)$ is defined as the predictive error, that is, ${\bf{e}}(k)=\bm{\varsigma}(k)-{{\bm{\varsigma}}_{\bf{L}}}(k)$, and ${{\bm{\varsigma}}_{\bf{R}}}(k)$ denotes the reference image feature without the predictive error. Thus, we can obtain:

\begin{equation}
\label{eq:referenceimage}
{{\bm{\varsigma}}_{\bf{R}}}(k)-{{\bm{\varsigma}}_{\bf{L}}}(k)={{\bm{\varsigma}}_{\bf{G}}}(k)-\bm{\varsigma}(k)
\end{equation}

The object of the visual MPC controller is then transformed into minimizing the tracking error of the prediction model with respect to ${{\bm{\varsigma}}_{\bf{R}}}(k)$. Let ${{\bm{\varsigma}}_{\bf{L}}}(k)={{\bm{\varsigma}}_{\bf{L}}}(k+1)-{{\bm{\varsigma}}_{\bf{L}}}(k)$, Eq. (\ref{eq:jacobian}) can be rewritten as the following state-space representation:
\begin{equation}
\label{eq:statespace}
\left\{\begin{matrix} 
{\bf{x}}(k+1)={\bf{x}}(k)+{\bf{B}}(k){\bf{u}}(k)\\
{\bf{y}}(k)={\bf{x}}(k)
\end{matrix}\right. 
\end{equation}
where the system state ${\bf{x}}(k)={{\bm{\varsigma}}_{\bf{L}}}(k)$, the control variable ${\bf{u}}(k)=\Delta{\bf{q}}(k)$, ${\bf{y}}(k)$ is the output and ${\bf{B}}(k)={\bf{\hat{L_m}}}(k){\bf{\hat{J}}}(k)$.

\subsection{Constraints}
In addition, some constraints should be considered. To ensure that the MicroNeuro remains stable and avoid undesirable contact with the brain ventricles, the camera position should meet certain constraint:
\begin{equation}
\label{eq:cameraconstraint}
{\bf{P}}^{\min}\leq{\bf{P}}(k)\leq{\bf{P}}^{\max}
\end{equation}

Correspondingly, considering some physical hard constraints on MicroNeuro, such as the restriction of the capability of the motors, actuator constraint is defined as follows:
\begin{equation}
\label{eq:actuatorconstraint}
{\bf{q}}^{\min}\leq{\bf{q}}(k)\leq{\bf{q}}^{\max}
\end{equation}

Moreover, to ensure that the target of concern is always within the field of view and away from areas with large camera distortion, output constrain is described as follows:
\begin{equation}
\label{eq:outputconstrain}
{{\bm{\varsigma}}_{\bf{L}}}^{\min}\leq{{\bm{\varsigma}}_{\bf{L}}}(k)\leq{{\bm{\varsigma}}_{\bf{L}}}^{\max}
\end{equation}
\subsection{Optimization Objective}
At each sample time $k$, the current measured system state is set as the initial state of an optimal control problem (OCP) with constrains, and the current control action is determined by solving the problem in the further $N_P$ sampling periods. Only the first optimal input is applied on the system in the optimal input sequence of length $N_c$. $N_p$ and $N_c$ are identified as the prediction horizon and control horizon, respectively. The objective is described as follows:
\begin{equation}
\label{eq:objective}
\begin{split}
&\min_{{\bf{U}}(k)}{\bf{V}}({\bf{U}}(k))=\sum_{i=k}^{k+N_p-1} ||{\bf{Y}}(k)-{\bf{S}}(k)||^2_{{\bf{Q}}}\\
&=\sum_{i=k}^{k+N_p-1} ({\bf{y}}(i|k)-{{\bm{\varsigma}}_{\bf{R}}}(i|k))^{\mathrm{T}}{\bf{Q}}({\bf{y}}(i|k)-{{\bm{\varsigma}}_{\bf{R}}}(i|k))
\end{split}
\end{equation}
subject to Eq. (\ref{eq:referenceimage}), (\ref{eq:cameraconstraint}), (\ref{eq:actuatorconstraint}) and (\ref{eq:outputconstrain}). In Eq. (\ref{eq:objective}), ${\bf{U}}(k)\in \mathbb{R}^{8N_p\times1}$, ${\bf{U}}(k)=({\bf{u}}(k|k)\cdots{\bf{u}}(k+N_c-1|k)\cdots{\bf{u}}(k+N_c-1|k))^{\mathrm{T}}$ is the control sequence, ${\bf{Y}}(k), {\bf{S}}(k)\in{\mathbb{R}}^{2N_p\times1}$ are output and reference sequence, ${\bf{Q}}\in\mathbb{R}^{2\times2}$ is the weight matrix. ${\bf{y}}(i|k)$ denotes the predictive value of output at $i$-th sample time. Problem (\ref{eq:objective}) can further come down to solve a quadratic programming (QP) with constrains. Specially, in our implementation, problem (\ref{eq:objective}) is formulated in CasADi \cite{Andersson2019} and is solved using its built-in optimization solvers.

\section{Experiment and Validation}
In this section,  we implemented four IBVS scenarios to evaluate the effectiveness of the proposed MicroNeuro robot and visual MPC controller. The  camera was well calibrated \cite{zhang2000flexible} with a low mean reprojection error of merely 0.2 pixels, and the image resolution was resized to $710\times710$ pixels from the origin resolution $400\times400$. This vision system was specifically designed to track the AprilTags \cite{olson2011tags}, which served as detection features and provided high accuracy localization. The tracking error in following was quantified as the Euclidean distance $\epsilon(k)$ between the measured and target coordinates of the features. In following experiments, the initial configuration of the robot is in steering mode 1 and remain straight. The kinematics was initialized with ${\bf{\hat{J}}}(0)$ and iterated online with $\epsilon(k)$. In the proposed visual MPC controller, the control horizon and prediction horizon are set to $N_c=N_p=10$, ${\bf{Q}}$ = diag$\left\{1,1\right\}$. According to \cite{vuong2021incidence}, the average tumor size in the pineal region is 26 mm. Based on Eq. (\ref{eq:pinhole}) and camera parameters, the Maximum Permissible Error (MPE) was defined  as $2.6$ mm,  and the corresponding pixel error is 30.

\begin{figure}[htbp]
  \vspace{3mm}
  \centerline{\includegraphics[width=1\linewidth]{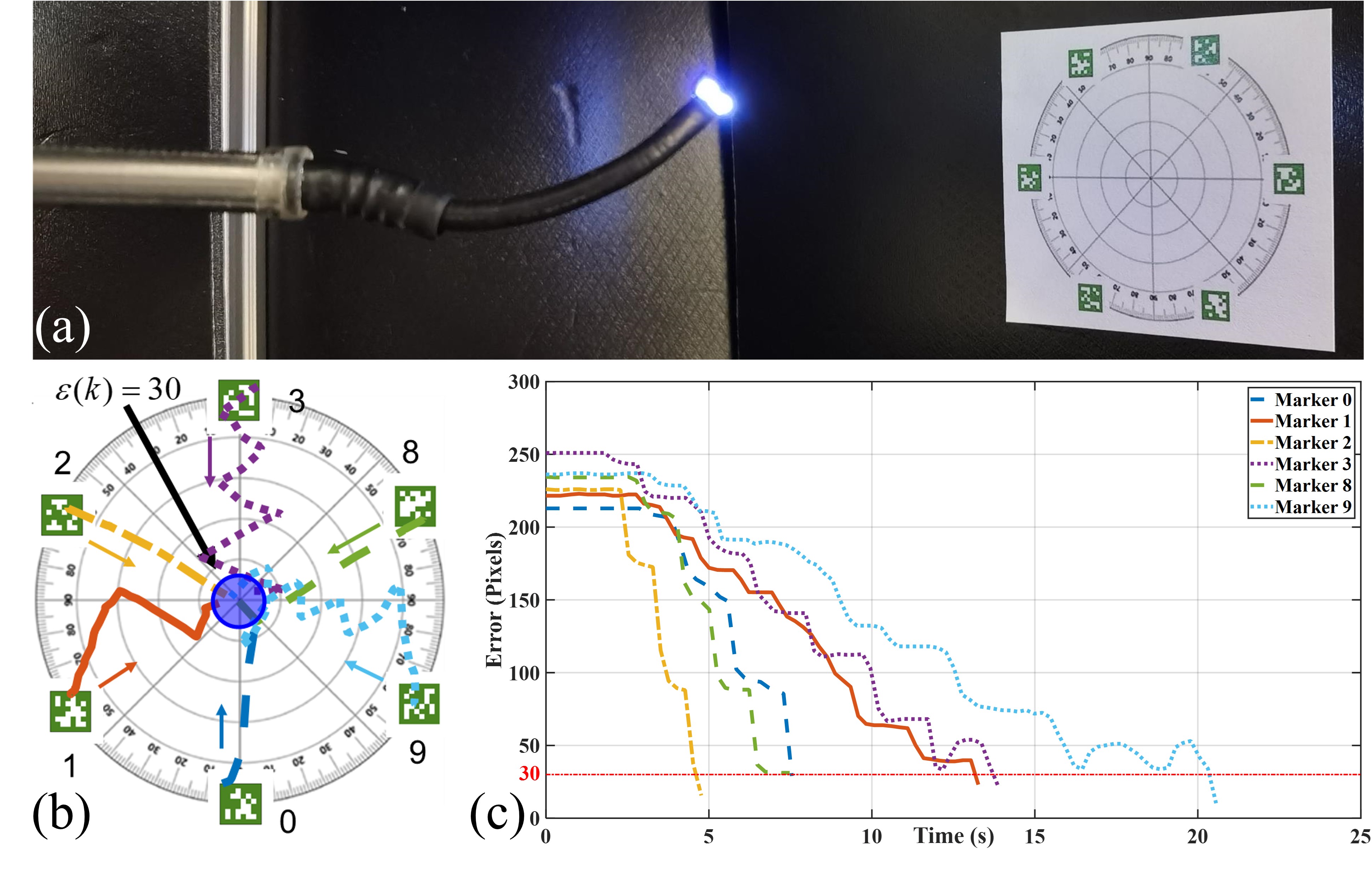}}
      \caption{Tracking the static object on a plane. (a) Experiment setup. (b) Tags movement trajectories in the image plane. (c) Tracking errors.}
      \label{fig5:static}
         \vspace{-2mm}
\end{figure}

\begin{figure}[htbp]
  \vspace{3mm}
  \centerline{\includegraphics[width=1\linewidth]{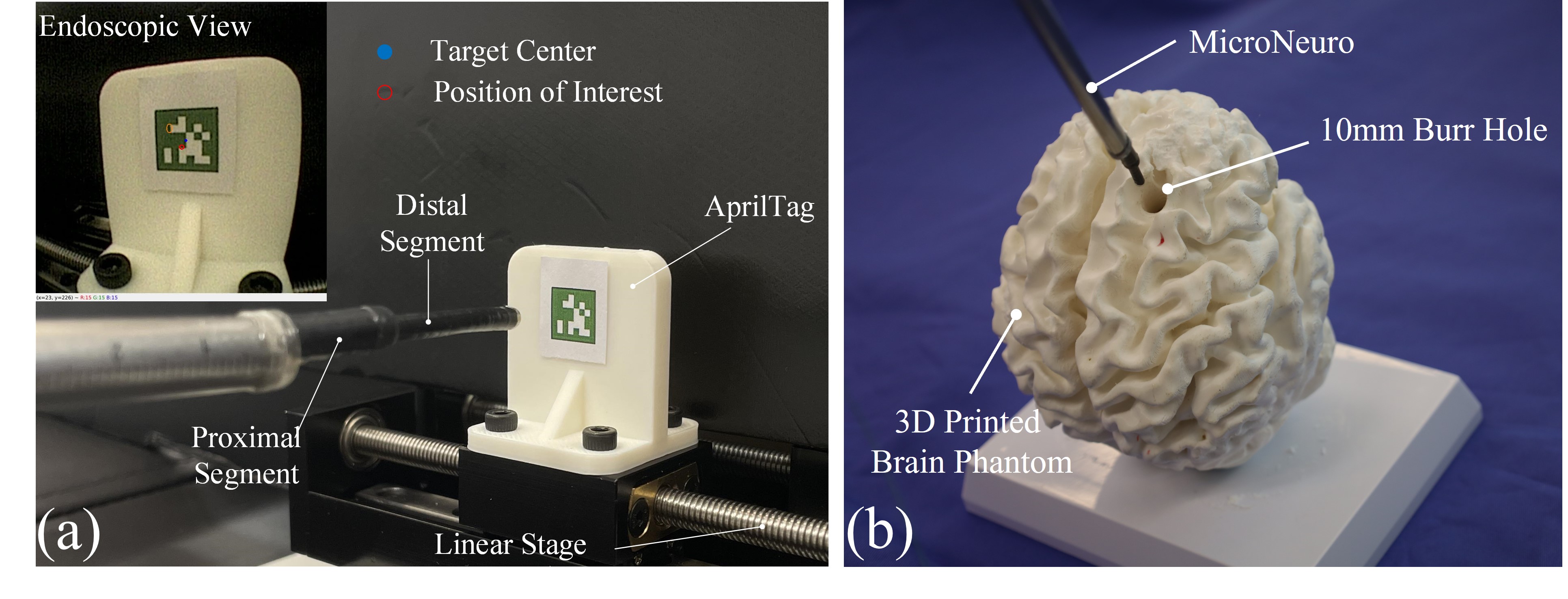}}
      \caption{Setups for: (a) Dynamic target tracking. (b) Biopsy in a brain phantom.}
      \label{fig6:setup}
         \vspace{-2mm}
\end{figure}

  \begin{figure}[!t]
  \vspace{3mm}
          \centerline{\includegraphics[width=1\linewidth]{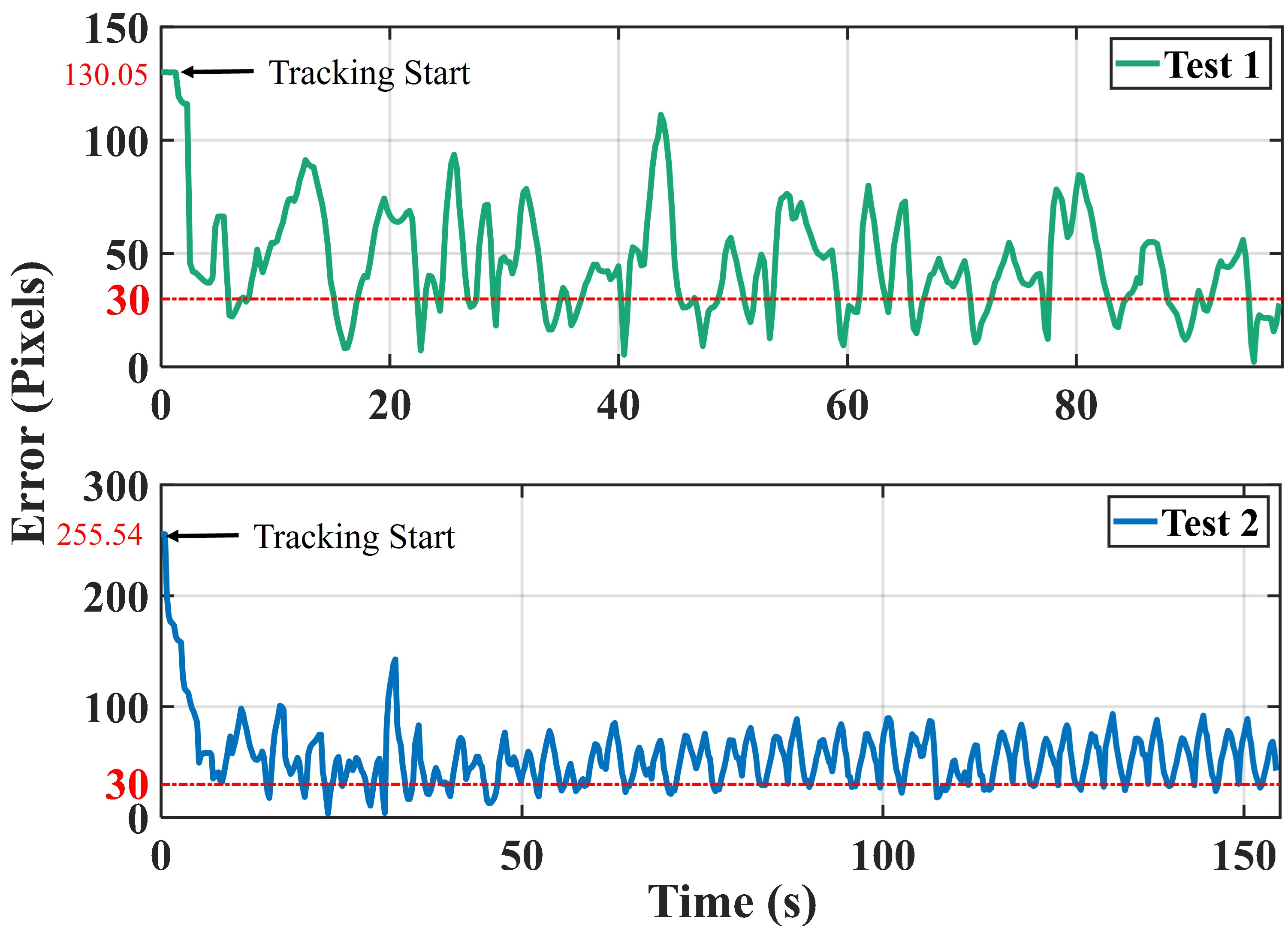}}
      \caption{Two test results of tracking the dynamic object on a linear guide.}
      \label{fig7:dynamic}
         \vspace{-2mm}
   \end{figure}

\begin{figure}[!b]
        \vspace{3mm}
\centerline{\includegraphics[width=1\linewidth]{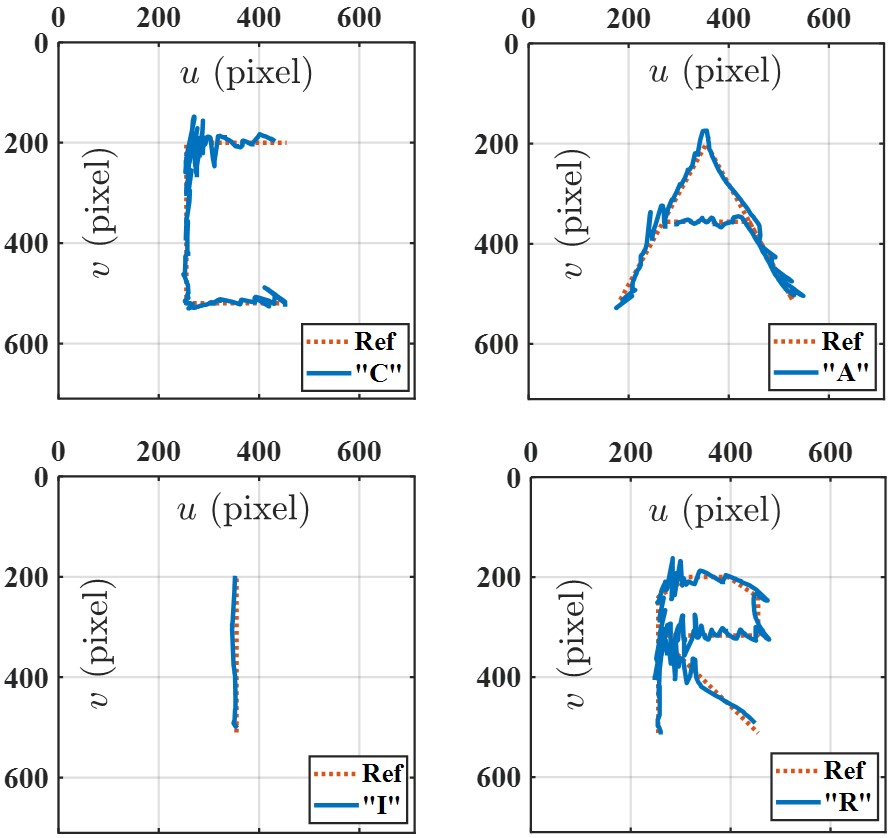}}
        \caption{"CAIR" Trajectories following results.}
        \label{fig:CAIR}
        \vspace{-2mm} 
\end{figure}

  \begin{figure*}[!t]
  \vspace{3mm}
      \centering
        \includegraphics[width=1\textwidth]{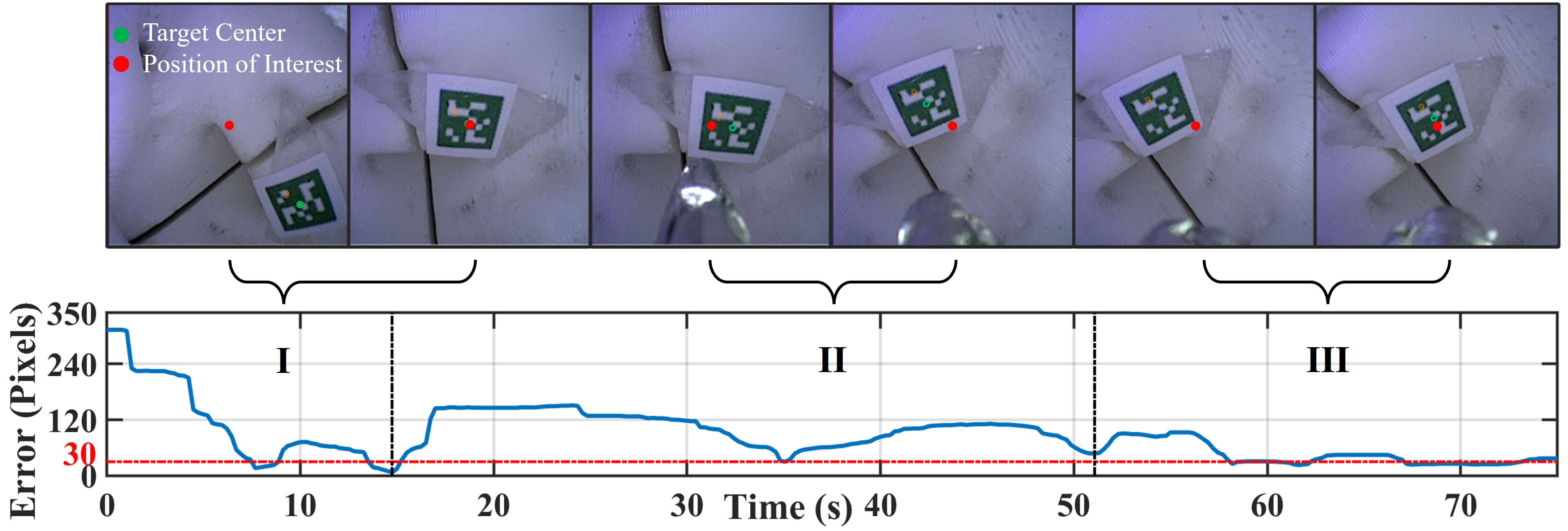}
      \caption{Various scenarios for target tracking in a 3D printed brain phantom.}
      \label{fig:real}
         \vspace{-4mm}
   \end{figure*}
\subsection{Static Target Tracking}
In this experiment, the region of interest (ROI) was defined as the center of the image ${{\bm{\varsigma}}_{\bf{G}}}(k)=(355,355)^{\mathrm{T}}$. As shown in Fig. \ref{fig5:static}(b), the MicroNeuro system was commanded to bring the specifically chosen markers to the ROI, which were distributed at $60^{\circ}$ intervals on a printed circle.  The experimental analysis involved conducting six trials, and the effectiveness was demonstrated through the measured trajectories of the markers, as depicted in Fig. \ref{fig5:static}(c). In each instance, the robot successfully returned the marker to the center with average terminal error was 21.8 pixels. The average time required to complete the tracking task across the six experiments was measured to be 11.25 s. This accomplishment highlights the robustness and reliability of the proposed method in achieving fast and precise tracking.

\subsection{Dynamic Target Tracking}
The experiment was designed to evaluate the stability of the proposed system following a target in a dynamic environment. As shown in Fig. \ref{fig6:setup}(a), the robot tracked an AprilTags marker attached to a linear guide, positioned 20mm from the robot's camera. The guide reciprocated at a speed of $2.5 mm/s$ over a $20 mm$ stroke. Fig. \ref{fig7:dynamic} illustrates that tracking errors decreased significantly once the marker was captured by the camera, with errors reduced to below the MPE within 6 s in Test 1, reaching a lowest error of 2.23 pixels. After the initial stable tracking of the target was accomplished, the standard deviation (SD) of the errors for test 1 and 2 were 20.85 and 21.81 pixels, respectively, which further supports the effectiveness of the system in maintaining precise tracking of the target. 

\subsection{Trajectory Following}

This experiment was designed to evaluate the robot's ability to follow a set trajectory that guides the marker along a defined path in the captured image. Experiment setup was same as Fig. \ref{fig5:static}(a). Under the guidance of the controller, the robot automatically completes tracking of multiple key target points on different trajectories to approximately complete the tracking of curves in the image plane. These discrete key target points set on the letters ${CAIR}$. The experimental results in Fig. \ref{fig:CAIR} showed that the controller has good tracking performance for the key points of each trajectory. The root mean square error (RMSE) of the four curves were 11.66, 11.62, 11.30 and 11.95 pixels respectively.

\subsection{Biopsy in a Brain Phantom}

In clinical procedures, the use of endoscopic instruments like biopsy gripper and electrocoagulation, inserted via the working channel, can significantly disrupt the flexible endoscope's view, leading to loss of lesion visibility or inadequate operating angles. This experiment aims to assess the robustness of the proposed method against external disturbances, ensuring the endoscope stays focused on the ROI. In the 3D printed brain shown in Fig. \ref{fig6:setup}(b), we placed a marker in the pineal gland region to mark the area of interest. Initially, the robot was manually operated to roughly approach the target area through one burr hole, and the visual MPC controller has quickly tracked the target, as shown in Fig. \ref{fig:real}. The insertion and operation of biopsy forceps introduced rapid noise to the robot, significantly increasing tracking error. However, the controller adjusted the tool within ten steps, reducing the error to less than 30 pixels. This result demonstrates the controller's ability to enhance the MicroNeuro robot's resistance to interference, suggesting its potential application in neurosurgery.

\section{Conclusion}
The presented study in this paper proposes a novel hybrid dual-segment flexible endoscope for neurosurgery. The dual-segment design allows for dexterous maneuverability within the deep brain's complex structure. This innovative approach substantially assists surgeons in performing procedures on the pineal region concurrently through a single burr hole, thereby enhancing surgical efficiency. The robot meets mechanical design requirements based on clinical needs and provides comprehensive endoscopic functionality. In addition, a visual servoing control system with online estimation of the Jacobian matrix is constructed to improve the motion performance of the robot. Considering unknown disturbance, a visual MPC with constraints has been designed. The experiment verified that the MicroNeuro robot is capable of executing precise visual servoing despite external interference, and demonstrated great potential for clinical applications in neurosurgery. In the future, this work will further consider the nonlinear dynamic model and the impact of contact force during intracranial surgery to enhance the performance of the visual  model predictive controller.

\section{Acknowledgements}
This work was supported by the Centre of AI and Robotics, Hong Kong Institute of Science and Innovation, Chinese Academy of Sciences, sponsored by InnoHK Funding, HKSAR, and partially supported by Sichuan Science and Technology Program (Grant number: 2023YFH0093). Parts of Fig. \ref{fig1}(a) were created using templates from Servier Medical Art (http://smart.servier.com/), licensed under a Creative Common Attribution 3.0 Generic License.

\bibliographystyle{IEEEtran}
\bibliography{root}
\addtolength{\textheight}{-12cm}   
\end{document}